\documentclass[10pt, a4paper]{article}
\usepackage[final]{lrec-coling2024} 

\usepackage{graphicx}
\usepackage{tabularx}
\usepackage{soul}
\usepackage{times}
\usepackage{latexsym}
\usepackage{hyperref}
\usepackage{url}
\usepackage{booktabs}
\usepackage{xcolor}
\usepackage{svg}
\usepackage{float}
\usepackage{natbib} 

\usepackage{titlesec}

\usepackage{epstopdf}
\usepackage[utf8]{inputenc}

\usepackage{hyperref}
\usepackage{xstring}

\usepackage{color}

\title{Exploring News Summarization and Enrichment in a Highly Resource-Scarce Indian Language: A Case Study of Mizo}

\name{Abhinaba Bala, Ashok Urlana, Rahul Mishra, Parameswari Krishnamurthy} 

\address{International Institute of Information Technology - Hyderabad \\
         \{abhinaba.bala, ashok.urlana, rahul.mishra, param.krishna\}@iiit.ac.in\\}

\abstract{
Obtaining sufficient information in one's mother tongue is crucial for satisfying the information needs of the users. While high-resource languages have abundant online resources, the situation is less than ideal for very low-resource languages. Moreover, the insufficient reporting of vital national and international events continues to be a worry, especially in languages with scarce resources, like \textbf{Mizo}. In this paper, we conduct a study to investigate the effectiveness of a simple methodology designed to generate a holistic summary for Mizo news articles, which leverages English-language news to supplement and enhance the information related to the corresponding news events. Furthermore, we make available 500 Mizo news articles and corresponding enriched holistic summaries. Human evaluation confirms that our approach significantly enhances the information coverage of Mizo news articles. The mizo dataset and code can be accessed at \url{https://github.com/barvin04/mizo_enrichment} . \\ 
\newline  }

\begin{document}

\maketitleabstract

\section{Introduction}
Low-resource languages often lack the required data resources for natural language processing tasks, hindering their inclusion in various applications. Significant progress has been made in generating open-source data for several scheduled Indian languages. However, languages like \textbf{Mizo} face persistent challenges in accessing domain-specific information. Mizo, a prominent member of the Tibeto-Burman language family, is primarily spoken by the Mizo people in India's northeastern region, especially in Mizoram, with significant populations in Manipur, Tripura, and Meghalaya. Additionally, it is also spoken in some parts of Myanmar and Bangladesh, further contributing to its linguistic diversity in the South Asian region. According to the 2011 census, the Mizo language had around 840,000 native speakers\footnote{\url{https://censusindia.gov.in/census.website/}}. Mizo uses the Roman alphabet for its script. 

Despite the presence of numerous newspapers in Mizo, limited NLP research has been conducted on this language. It is important to note that the sheer existence of numerous newspapers does not necessarily translate into an abundance of resources suitable for training NLP models. The scarcity of data in such languages remains a significant obstacle to performing essential NLP tasks, despite the progress made in this field for other languages.
\begin{table}[t]
    \centering\small
    \begin{tabularx}{\linewidth}{X}
        \toprule
        \textbf{Mizo text (truncated):} Nimin khan Saron Veng, \textcolor{blue}{Serchhip district} atangin chhungkaw 7 chu sawnchhuah an ni a, nimin khan \textcolor{blue}{ruahsur nasa avangin}... nimin khan sawnchhuah an ni National Highway 54...occurred last year Chhungkaw pariat an awm tawh an chenna in aṭanga chhuahtiran ni. \\
        \textbf{En\_translated:} Seven families from Saron Veng, \textcolor{blue}{Serchhip district} were evacuated yesterday  and \textcolor{blue}{today due to heavy rains} ...were evacuated today National Highway 54...occurred last year Eight families have been evacuated from their homes. \\
        \midrule
        \textbf{En\_enriched (truncated) :} \textcolor{magenta}{Eight people have been killed and six are missing} after \textcolor{blue}{flash floods} caused by \textcolor{blue}{heavy rainfall} wrecked havoc in Tlabung in Mizoram’s \textcolor{blue}{Lunglei district}.... \textcolor{magenta}{350 houses have been submerged since yesterday.} \\
        \textbf{Mizo\_translated:} Mizoram’s \textcolor{blue}{Lunglei district-a} Tlabung khuaah \textcolor{blue}{ruahsur nasa vanga} tuilianin a tihchhiat avangin mi \textcolor{magenta}{8 an thi tawh a}, mi paruk chin hriat lohin an awm tawh a.... \textcolor{magenta}{Nimin aṭang khan in 350 tuiin a chim tawh a ni.}\\
        \bottomrule
    \end{tabularx}
    \caption{Example of the \textbf{(Top)} raw Mizo news article and corresponding translated version.  \textbf{(Bottom)} corresponding enriched version of the same. Highlighted in \textcolor{magenta}{\textbf{magenta}} indicates the enrichment part, whereas \textcolor{blue}{\textbf{blue}} signifies the context of the original article.}
    \label{tab:mizo_output}
\end{table}


The process of enriching articles written in low-resource languages through the utilization of auxiliary information represents an important advancement in the field of natural language processing. This auxiliary information serves as a repository of more pertinent and coherent information related to the original Mizo text as shown in Table~\ref{tab:mizo_output}. By incorporating auxiliary information, which could include translations, named entity recognition, summarization, information extractions, transcriptions, or contextual data from more widely studied languages, the Mizo articles gain depth, clarity, and broader accessibility. This approach not only enhances the overall quality of content but also contributes to the continued documentation and dissemination of languages that might otherwise face the risk of being marginalized or lost over time. The utilization of auxiliary information exemplifies the intersection of technology and language conservation, fostering a bridge between underrepresented languages and the digital age while reinforcing the importance of linguistic diversity.

Our pipeline does not assume that events covered in Mizo and English news media are almost parallel. Instead, it aims to enrich the original Mizo articles with additional information available in English when feasible. The goal is to supplement the content, but we acknowledge that English may not always contain extra information on the specific events covered in the Mizo dataset.

In this study, we aim to enrich articles in low-resource languages by supplementing them with relevant information extracted from high-resource languages, such as English, using state-of-the-art NLP techniques. We introduce a straightforward pipeline that includes the following steps:

\begin{itemize}
    \item Translate Mizo news article into English and generating a headline using state-of-the-art headline generation models.
    \item Extract valid URLs by querying the generated headline in a web search.
    \item Retrieve documents from the identified URLs and perform the multi-document summarization using state-of-the-art pre-trained models.
    \item Add the obtained summary to the corresponding document and translate the entire English document back to the Mizo language. 
\end{itemize}
We have released the 500 Mizo documents and their corresponding enriched versions to facilitate further research on low-resource languages. To assess the pipeline's performance, we conducted a human evaluation. The results of the human evaluation indicate that the proposed pipeline effectively enriches low-resource language news articles. 

\section{Related Work}

\subsection{Mizo Datasets}
Comprehensive datasets for Mizo language tasks are scarce, with a predominant focus on fundamental language understanding and translation rather than the creation of holistic summaries of Mizo news articles. Notably, research efforts such as \citet{10028882} aim to address the scarcity of multimodal datasets for low-resource language pairs like English-Mizo. They present the Mizo Visual Genome 1.0 (MVG 1.0) dataset, featuring bilingual textual descriptions alongside images, facilitating English-Mizo multimodal machine translation.
Additionally, \citet{lus} contributes an LUS dataset, a collection of 101,827 monolingual Mizo language sentences sourced from various news websites. 

\subsection{Headline Generation}
Generating headlines \cite{Zhou2004TemplateFilteredHS, Alotaiby2011AutomaticHG, PanthaplackelETAL22UpdatedHeadlineGeneration} from articles simplifies information access and exploration. These succinct headlines can serve as search queries, aiding users in finding more related articles efficiently \citep{webSearch2015}.

\subsection{Multi Document Summarization}
Multi-document Summarization (MDS) involves the generation of a brief and condensed summary that includes the essential information from a collection of interconnected documents. 
Recent studies in MDS have demonstrated promise in both extractive \cite{angelidis-lapata-2018-summarizing, narayan-etal-2018-ranking} and abstractive \cite{Chu2018MeanSumAN, fabbri-etal-2019-multi, liu-lapata-2019-hierarchical} summarization techniques.




\begin{table}[t]
\centering
\begin{tabular}{l|l}
\toprule
\textbf{Description} & \textbf{Count} \\
\midrule
Mizo (single news) Documents & 983 \\
Mizo translated to English & 983 \\
Headlines & 798 \\
Articles with (valid + invalid) URLs & 797 \\
Articles without URLs & 30 \\
Articles with valid URLs ($\geq 1$) & 767 \\
Articles with valid URLs ($\geq 2$) & 746 \\
Total URLs & 4054 \\ \midrule
Average URLs per document & \textbf{5.29}\\
\bottomrule
\end{tabular}
\caption{Mizo data statistics}
\label{tab:data_stats}
\end{table}

\begin{figure*}
  \centering
  \includegraphics[width=0.8\textwidth]{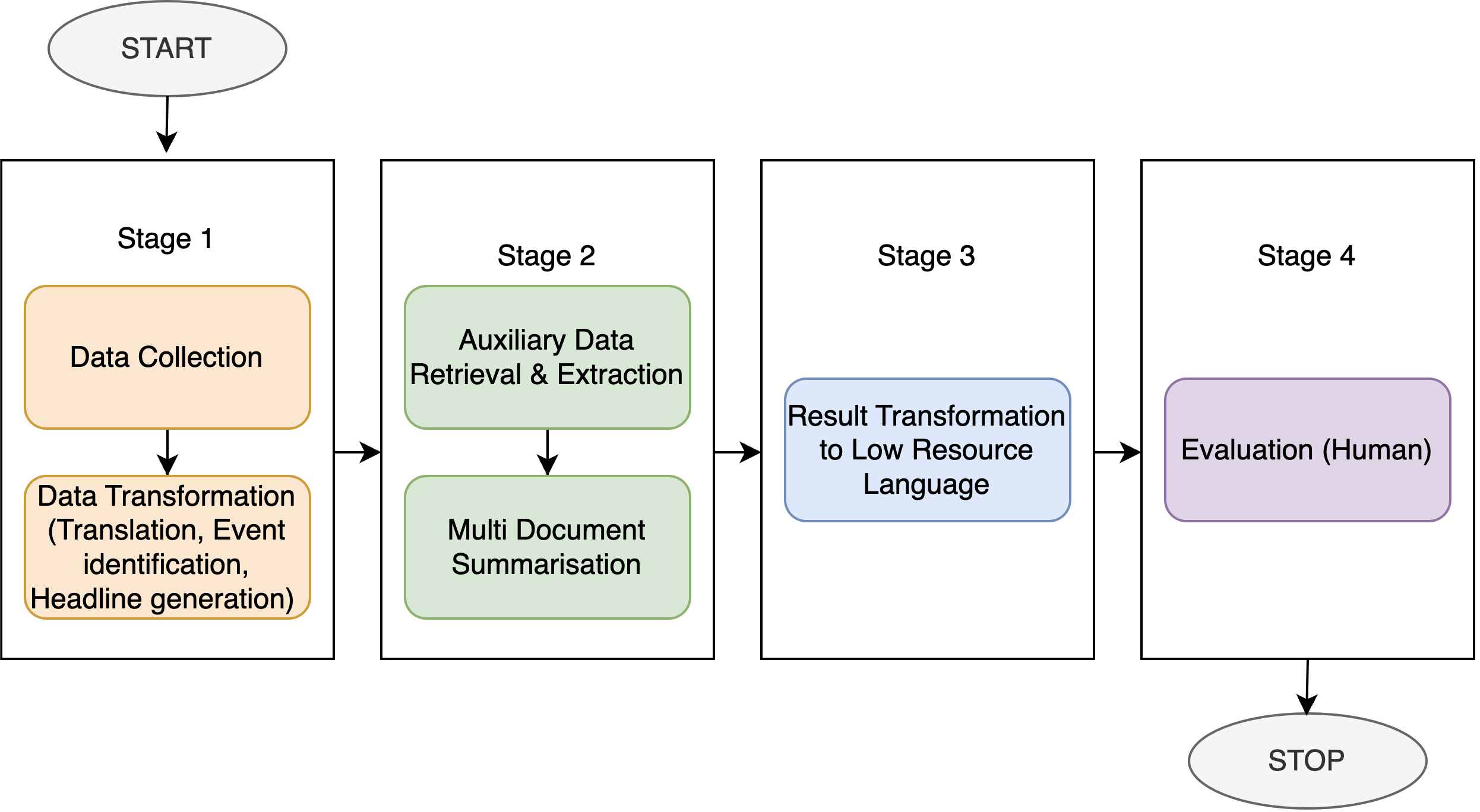}
  \caption{The Enrichment Methodology Pipeline. This illustration outlines the sequential stages of the methodology, which encompasses \textbf{a.} data collection, \textbf{b.} preprocessing, transformation/translation, \textbf{c.} headline generation, \textbf{d.} multi-document summarization, and \textbf{e.} translation into the low-resource language. These stages collectively contribute to the enrichment of articles in low-resource languages, facilitating a comprehensive understanding and accessibility of the content.}
  \label{fig:method overview}
\end{figure*}

\section{Methodology}
The methodology of this work leverages a simple pipeline that leverages state-of-the-art natural language processing (NLP) techniques to enrich articles in low-resource languages, such as Mizo, through the incorporation of auxiliary information from high-resource languages like English. The overarching process can be delineated into several key stages as shown in Figure~\ref{fig:method overview}.

    

\subsection{Data acquisition and preprocessing}
To obtain Mizo raw data, we scraped publicly available information exclusively from the Mizoarchive\footnote{\url{https://mizoarchive.wordpress.com/}}, an online news portal. To ensure the creation of a high-quality dataset, we subject the collected data to necessary rule-based preprocessing steps. This involves the elimination of HTML tags and the removal of noisy text elements to preserve the integrity and quality of the source documents.

\subsection{Data transformation}
 The data transformation includes translation from Mizo to English and obtaining the headline from the corresponding Mizo documents. 

\noindent \textbf{Translation from Mizo-English: } Due to the absence of a Mizo summarization model, the initial step in our pipeline involves translating the cleaned Mizo document into English. We have utilized the Google-translate API to obtain Mizo to English translation. For the upcoming stages in the pipeline we have utilized English translated Mizo document. 

\noindent \textbf{Headline generation}
We employed state-of-the-art headline generation models to create headlines from English-translated Mizo documents. Specifically, we use the BART-large model \cite{lewis-etal-2020-bart} fine-tuned on the CNN dataset for the headline generation task. 
\subsection{Information extraction}
\noindent \textbf{Obtaining valid URLs:} 
Upon querying the headline in a web search, we retrieved various URLs. 
A URL was deemed valid if it directed to a Mizo news-article. 
We excluded URLs from major platforms like Wikipedia and YouTube. Comprehensive details on the criteria defining a valid URL are available in Table~\ref{tab:data_stats}. Our approach encompasses documents of varying lengths, and we consider all topics without selective exclusions, ensuring a thorough exploration. On average, we obtained 5.29 valid URLs for each query.

\noindent \textbf{Information retrieval: } To acquire pertinent information from each web page linked to valid URLs, we employed a web scraping technique to transform unstructured web data into a structured format. This process involved utilizing the "Google" search engine and Python libraries like BeautifulSoup and urllib2. Specifically, urllib2 was used for URL retrieval, while BeautifulSoup was employed for data extraction.

In the pursuit of data quality and relevance, we meticulously selected the most contextually relevant URLs for the query. We also extracted valuable meta information from HTML tags like headings, paragraphs, tables, and images. Any incomplete sentences or irrelevant headings were intentionally excluded. After careful consideration of multiple sources, meaningful sentences were extracted and consolidated into a single document.

\subsection{Uni-document Summarization}
After the information extraction step, for each En-Mizo document, we have more than one relevant document. The assumption would be each document covers the relevant information with respect to the En-Mizo document. In this step, we have utilized the PEGASUS \cite{zhang2020pegasus} large model to generate individual summaries for each document. Subsequently, we concatenated all these summaries and fed the concatenated result back into the PEGASUS model to produce a coherent summary of all the documents.


\subsection{Enrichment of Low Resource Language Articles}
The final step of this pipeline is to translate all the English summaries into Mizo. This step ensures the conversion of the outcome from the high-resource language(s) into the target low-resource language. This step is pivotal in ensuring that the conclusions, findings, and insights derived from the analysis are made accessible and comprehensible to users who primarily operate within the context of the low-resource language. The next section validates the quality of the corresponding summary by performing the human evaluation. 

\section{Human Evaluation}
\subsection{Guidelines}
To assess the quality of the pertinent information acquired through the proposed methodology, we conducted human evaluation. We randomly selected 50 documents from our meticulously curated Mizo dataset and engaged a native and proficient Mizo speaker to evaluate the generated summaries. We have provided the original Mizo document and the obtained summaries from the proposed pipeline and instructed the evaluator to assess the quality based on the following four distinct categories:
\begin{itemize}
    \item Coherency: Assessing the logical flow and consistency of the summaries. 
    \item Enrichment: Evaluating how effectively the summaries enhanced the original content.
    \item Relevancy: Determining the degree of relevance of the summaries to the original documents.
    \item Readability: Gauging the ease with which the summaries could be comprehended.
\end{itemize}
Each category was assessed on a scale from 0 to 4, with 0 indicating very poor, 1 representing somewhat unfaithful, 2 denoting moderate, 3 indicating good, and 4 representing near-perfect performance.

\subsection{Analysis and discussion}

\begin{table}
\centering\scriptsize
\begin{tabular}{l|c|c|r}
\toprule
\textbf{Coherency} & \textbf{Enrichment} & \textbf{Relevancy} & \textbf{Readability} \\
\midrule
3.82 & 2.44 & 2.9 & 3.98 \\
\bottomrule
\end{tabular}
\caption{Human evaluation results}
\end{table}

\begin{itemize}
    \item Coherency: The evaluation resulted in a relatively high level of coherency (3.82), suggesting that the logical flow and consistency of the summaries are generally well-maintained, contributing to their overall quality.
    \item Enrichment: The enrichment category received a score of 2.44, indicating moderate effectiveness in enhancing the original content within the generated summaries. The summaries appear to contribute to enriching the original content to a reasonable extent, but there are opportunities for refinement to make them more effective in this regard.
    \item Relevancy: The obtained score for relevancy is 2.9, suggesting a moderately relevant connection between the summaries and the original documents. This score indicates that the summaries exhibit a degree of alignment with the source documents, providing a basis for understanding the content. 
    \item Readability: The readability score was relatively high at 3.98, indicating that the summaries are generally easy to comprehend. This is a positive aspect, as it ensures that the information can be accessible to a broader audience.
\end{itemize}

While coherency and readability seem to be relatively strong points, there is room for improvement in terms of enrichment and relevancy. These findings can guide future refinements of the summary generation pipeline, with the aim of achieving more comprehensive and contextually relevant summaries that enhance the original content to a greater extent.
\section{Conclusion}

In this work, we introduced a simple pipeline for enhancing low-resource (Mizo) news articles by infusing them with contextually relevant information. Our approach significantly improves the coverage of pertinent topics within Mizo documents, which is apparent from the results of our human evaluation. Additionally, this pipeline can be utilized to boost news content in other underrepresented languages with only minor modifications to the overall approach.

\section{Limitations}
The effectiveness of the proposed methodology relies on the availability and relevance of auxiliary information from high-resource languages. In scenarios where such information is sparse or not applicable, the enrichment process may be hindered. The assumption that events covered in Mizo and English news media are parallel may not always hold true. Variations in news coverage and the uniqueness of local events may challenge the assumption that English can consistently supplement Mizo articles.

The chosen evaluation metrics, while providing valuable insights, might have limitations in fully capturing the nuanced aspects of enriching low-resource language articles. Further refinement and exploration of evaluation methodologies could enhance the robustness of the assessments. Human evaluation, while insightful, is inherently subjective. The interpretation of coherency, enrichment, relevancy, and readability can vary among evaluators, introducing a level of subjectivity that might impact the reliability of the assessments.

\section{Ethics Statement}
In conducting this research, we have prioritized key ethical considerations to uphold the integrity and responsibility of our work. Data privacy and informed consent are important, particularly when involving human subjects, ensuring that personal information is treated confidentially. 
Transparency is maintained through clear disclosure of data sources, methodologies, and any limitations present in the study. Cultural sensitivity is observed, avoiding misrepresentation and respecting the diversity of communities involved. Embracing open science practices, we aim to share code, datasets, and findings openly to foster collaboration and reproducibility. 

\section*{Acknowledgements}
We would like to express our sincere gratitude to the Mizo annotator(s). 

\section*{References}
\bibliographystyle{lrec-coling2024-natbib}
\bibliography{lrec-coling2024-example}
\newpage
\appendix
\section{Appendix : Examples}
\label{sec:appendix}

As shown in Table ~\ref{tab:appendix_1}, the enriched summary obtain high scores (all 4) by human evaluation. Where the context of \textit{Turkey and US} is carried over as `The two NATO allies'. The additional enrichment by our pipeline adds in the information about \textit{equipment related to \textit{F-35 flighter aircraft}. }



Table ~\ref{tab:appendix_2} shows the text for an example with scores (4, 2, 3, 4) for coherency, enrichment, relevancy and readability respectively. While there is moderate enrichment, there are parts which are not relevant or enriching enough according to the annotator. 
\begin{table}[htb]
    \centering\small
    \begin{tabularx}{\linewidth}{X}
        \toprule
        \textbf{Mizo text (truncated):} \textcolor{blue}{Turkey} President Recep Tayyip Erdogan chuan \textcolor{blue}{US}-in Patriot missile a pawmpui chung pawhin Turkey chuan \textcolor{blue}{Russian S-400 missile} lei tumna chu a thulh dawn lo tih a sawi. ... \\
        \textbf{En\_translated:} \textcolor{blue}{Turkish} President Recep Tayyip Erdogan has said that Turkey will not cancel its plan to buy \textcolor{blue}{Russian S-400 missiles} despite the \textcolor{blue}{US} approval of Patriot missiles. ... \\
        \midrule
        \textbf{En\_enriched (truncated) :} ... \textcolor{blue}{The two NATO allies} have sparred publicly for months over Turkey's order for \textcolor{blue}{Russia's S-400 air defense system}, which Washington says poses a threat to the Lockhead Martin Corp F-35 stealthy fighters, which Turkey also plans to buy. The United States has \textcolor{magenta}{halted delivery of equipment related to the stealthy F-35 fighter aircraft} ...  \\
        \textbf{Mizo\_translated:}... \textcolor{blue}{NATO tangrual} ram pahnihte hi Turkey-in \textcolor{blue}{Russia-a S-400 air defense system} a order chungchangah thla tam tak chhung vantlang hmaah an inhnial tawh a, Washington chuan Lockheed Martin Corp F-35 stealthy fighter-te tan hlauhawn tak a nih thu a sawi a, Turkey pawhin lei a tum bawk. US \textcolor{magenta}{chuan F-35 fighter aircraft rukbo nena inzawm hmanrua pekchhuah chu a titawp} ... \\
        \bottomrule
    \end{tabularx}
    \caption{Appendix-1, example of the \textbf{(Top)} raw Mizo news article and corresponding translated version.  \textbf{(Bottom)} corresponding enriched version of the same. Highlighted in \textcolor{magenta}{\textbf{magenta}} indicates the enrichment part, whereas \textcolor{blue}{\textbf{blue}} signifies the context of the original article.}
    \label{tab:appendix_1}
\end{table}
    
\begin{table}[t]
    \centering\small
    \begin{tabularx}{\linewidth}{X}
        \toprule
        \textbf{Mizo text (truncated):} ... "Kan zinkawngah hian MNF sorkarin hun rei lote chhungin hma a sawn a. \textcolor{blue}{Chief Minister Pu Zoramthanga} chuan sorkar tharin hma a la dawn tih sawiin, conduct rules hian hun rei tak chhung min phuar tawh a ni" a ti a. \\
        \textbf{En\_translated:} ... "MNF government has made progress in our journey within a short period of time. \textcolor{blue}{Chief Minister Pu Zoramthanga} said that the new government is about to take action and the conduct rules have been binding us for a long time" ... \\
        \midrule
        \textbf{En\_enriched (truncated) :} Mizoram Chief Minister Zoramthanga on Sunday asserted that his party, the Mizo National Front, will return to power and bag 25-35 seats ... Mizoram \textcolor{blue}{CM Zoramthanga} Reacting to \textcolor{magenta}{allegations that the party is afraid of the BJP}, the CM said the BJP-led central government ...  \\
        \textbf{Mizo\_translated:} Mizoram \textcolor{blue}{Chief Minister Zoramthanga} chuan Pathianni khan a party, Mizo National Front chu thuneihna chang lehin seat 25-35 an la dawn tih a nemnghet a ... Mizoram CM Zoramthanga \textcolor{magenta}{BJP an hlauhthawn nia puhna chhangin}, CM chuan BJP kaihhruai central chu a sawi sawrkar ... \\
        \bottomrule
    \end{tabularx}
    \caption{Appendix-2, example of the \textbf{(Top)} raw Mizo news article and corresponding translated version.  \textbf{(Bottom)} corresponding enriched version of the same. Highlighted in \textcolor{magenta}{\textbf{magenta}} indicates the enrichment part, whereas \textcolor{blue}{\textbf{blue}} signifies the context of the original article.}
    \label{tab:appendix_2}
\end{table}


\end{document}